%% file: main.tex
\documentclass{article}

\usepackage[utf8]{inputenc} %
\usepackage[T1]{fontenc}    %
\usepackage{hyperref}       %
\usepackage{url}            %
\usepackage{booktabs}       %
\usepackage{amsfonts}       %
\usepackage{nicefrac}       %
\usepackage{microtype}      %
\usepackage{xcolor}         %

\usepackage{amsmath}
\usepackage{amssymb}
\usepackage{amsthm}
\usepackage{bm}
\usepackage{graphicx}
\usepackage{mathtools}
\usepackage{subfigure}
\usepackage[capitalize,noabbrev]{cleveref}
\usepackage{textcomp} %
\usepackage{minitoc}
\usepackage{authblk}

\usepackage[numbers]{natbib}
\theoremstyle{plain}
\newtheorem{theorem}{Theorem}[section]

\newtheorem{corollary}[theorem]{Corollary}
\theoremstyle{definition}
\newtheorem{definition}[theorem]{Definition}

\theoremstyle{remark}
\newtheorem{remark}[theorem]{Remark}
\newcommand{\Conv}{\mathop{\scalebox{1.7}{\raisebox{-0.2ex}{$\ast$}}}}%

\title{Improve Long-term Memory Learning Through Rescaling the Error Temporally}

\author[1]{Shida Wang \thanks{Corresponding author: \url{shida_wang@u.nus.edu}}}
\author[2]{ZhangLu Yan}
\affil[1]{Department of Mathematics, National University of Singapore}
\affil[2]{School of computing, National University of Singapore}

\begin{document}

\maketitle

\begin{abstract}
This paper studies the error metric selection for long-term memory learning in sequence modelling. 
We examine the bias towards short-term memory in commonly used errors, including mean absolute/squared error. 
Our findings show that all temporally positive-weighted errors are biased towards short-term memory in learning linear functionals. 
To reduce this bias and improve long-term memory learning, we propose the use of a temporally rescaled error. 
In addition to reducing the bias towards short-term memory, this approach can also alleviate the vanishing gradient issue. 
We conduct numerical experiments on different long-memory tasks and sequence models to validate our claims. 
Numerical results confirm the importance of appropriate temporally rescaled error for effective long-term memory learning.
To the best of our knowledge, this is the first work that quantitatively analyzes different errors' memory bias towards short-term memory in sequence modelling. 
\end{abstract}

\section{Introduction}

The challenge of comprehending long-term relationships is a perennial issue within the realm of sequence modeling. 
Sequences that manifest these characteristics are often classified as having a long ``memory''. 
The acquisition of knowledge pertaining to this prolonged memory is pivotal for various applications, notably in the field of time series prediction \citep{connor1994.RecurrentNeuralNetworksa}, machine translation \citep{wu2016.GoogleNeuralMachine}, speech recognition \citep{graves2013.SpeechRecognitionDeep}, language modelling \citep{paperno2016.LAMBADADatasetWorda}, reinforcement learning \citep{kirkpatrick2017.OvercomingCatastrophicForgetting}. 
To further improve the performance on these tasks, designing more scalable models \citep{cho2014.LearningPhraseRepresentations} and finding better optimization methods \citep{informatik2003.GradientFlowRecurrent} are two common directions.

Numerous models aimed at enhancing the retention of long-term memory in sequential data.
Recurrent-type neural networks such as RNN \citep{rumelhart1986.LearningRepresentationsBackpropagating}, GRU \citep{cho2014.LearningPhraseRepresentations} and LSTM \citep{hochreiter1997.LongShorttermMemory} are popular models for learning sequential relationship.
However, recurrent models can suffer from the vanishing gradient issue and they have an exponentially decaying memory \citep{pascanu2013.DifficultyTrainingRecurrent, li2022.ApproximationOptimizationTheory, wang2023.InverseApproximationTheory}. 
The ReLU activation was initially introduced to tackle the vanishing gradient problem prevalent in sequence modelling, as noted by \citet{pascanu2013.DifficultyTrainingRecurrent}.
More recently, innovative structures for sequence modeling have emerged, such as Temporal Convolution Networks (TCN) \citep{bai2018.EmpiricalEvaluationGeneric} and attention-based models \citep{vaswani2017.AttentionAllYou}.
These are attempts from the model construction perspective.

Apart from the model design aspect, various parameterization techniques have been explored.
Models like AntisymmetricRNN \citep{chang2018.AntisymmetricRNNDynamicalSystem}, UnitaryRNN \citep{arjovsky2016.UnitaryEvolutionRecurrenta}, IndRNN \citep{li2018.IndependentlyRecurrentNeuralb}, KRU\citep{jose2018.KroneckerRecurrentUnits}, ExpRNN \citep{lezcano-casado2019.CheapOrthogonalConstraintsa}, LEM\citep{rusch2022.LongExpressiveMemorya}, CoRNN \citep{rusch2022.CoupledOscillatoryRecurrent}, Hippo \citep{gu2020.HiPPORecurrentMemorya,smith2023.SimplifiedStateSpace} all can be regarded as the approach to seek better parameterization methods. 
While these models differ from the original recurrent networks, they do not expand the approximation space. 
Consequently, the advantages they offer in terms of speed and stability primarily stem from an optimization standpoint.
Besides the recurrent neural networks, \citet{romero2022.CKConvContinuousKernela} also proposes implicit parameterization for convolutional kernels which facilitate the learning of long memory. 

In this paper, we investigate the feasibility of learning long-term memory based on the error metric selection. 
We emphasize this approach is orthogonal to the previous approaches including model construction and parameterization selection. 
We evaluate the bias of error towards short-term memory. 
It is proved that commonly-used error such as mean absolute/squared error are biased towards short-term memory. 

To summarize, our main contibutions are 
\begin{enumerate}
    \item We identify the existence of bias towards short-term memory among mean absolute/squared error. In particular, we quantify the scale of memory bias in the linear functional case.  
    \item We extend the memory bias analysis to more general temporally positive-weighted errors. 
    \item We confirm the theoretical claims based on numerical experiments on linear functional learning, copying problem and text summarization task. 
    The effectiveness of rescaling error temporally are justified via different sequence models. 
\end{enumerate}

\paragraph{Paper structure}
In \cref{sec:problem_formulation}, we give the setup of the sequence modelling and the definition of memory function for general sequence to sequence relationship.
In \cref{sec:main_results}, we present the main results on the memory bias of MSE/MAE and general temporally positive-weighted errors. 
A subfamily of temporally positive-weighted errors are selected to compare the memory bias effects. 
In \cref{sec:numerical_results}, the numerical experiments are shown to validate the claims in \cref{sec:main_results}.
The related works from sequence modelling are summarized in \cref{sec:related_work}.

\section{Problem formulation}
\label{sec:problem_formulation}

We start with the setup of the sequence modelling. 
Next we summarize different memory definitions in sequence modelling. 
The memory evaluation method is important for the fair comparison of the memory bias. 

\subsection{Sequence modelling}

In sequence modelling, a map $\{H_t\}_{t \in \mathbb{R}}$ is learned between input sequence $\mathbf{x} = \{x_t\}$ and the corresponding output sequence $\mathbf{y} = \{y_t\}$.
This mapping can be regarded as sequence of functionals from the mathematical perspective. 
\begin{equation}
    y_t = H_t(\mathbf{x}).
\end{equation}
Notice here $t$ is not limited to discrete index. 
For theoretical study perspective, it's common to assume the index $t \in \mathbb{R}$ to be unbounded.
In this paper, when we consider the specific memory bias, we will focus on a fixed time horizon $t \in [0, T], T > 0$.
We make the additional assumption that the target sequential relationship is time-homogeneous. This characteristic is crucial for ensuring the effectiveness of a model when trained at one time interval and evaluated at a different time interval.

\subsection{Memory evaluation in sequence modelling}

Now we summarize different memory function definitions commonly used in sequence modelling and give the memory function definition used in our paper.

\paragraph{Finite horizon memory function}
Given sequences $\{x_t\}$ and $\{y_t\}$. 
Assume there exists a positive horizon $\tau$ such that the output $y_t$ is only a deterministic function of inputs $x_{t-\tau}, x_{t-\tau+1}, \dots, x_{t-1}, x_t$. 
\begin{equation}
    y_t = f(x_{t-\tau}, x_{t-\tau+1}, \dots, x_{t-1}, x_t).
\end{equation}
If there is no such positive horizon, it is said to have a memory with infinite horizon. 

\paragraph{Representation induced memory function}
Above finite horizon definition is not generalized enough as it only characterizes the length of memory without describing the decay property.
Exponential moving average (EMA) $y_t = \alpha y_{t-1} + (1-\alpha) x_t, \alpha \in (0, 1)$ all have memory with infinite horizon but the memory of the past inputs can decay differently. 
\begin{equation}
    \textrm{EMA: } y_t = \sum_{\tau=-\infty}^0 (1-\alpha) \alpha^{-\tau} x_{t+\tau}.
\end{equation}
The corresponding continuous version of the exponential moving average is
\begin{equation}
    \textrm{EMA (Continuous): } y_t = \int_{-\infty}^0 \alpha^{-s} x_{t+s} ds.
\end{equation}

For continuous, linear, time-homogeneous functionals with suitable properties (see the definitions and related representation results from \cref{appendix:linear_functional}), they have the following explicit representation and $\rho$ is a naturally induced memory function for functional sequence $\{H_t\}_{t \in \mathbb{R}}$
\begin{equation}\label{eq:rho_in_linear_functional}
    y_t = H_t(\mathbf{x}) = \int_{-\infty}^t \rho_{t-s} x_s ds.
\end{equation}
It can be seen the exponential moving average corresponds to the linear functional with memory function $\rho(t) = \alpha^t, t \in [0, \infty)]$.
A natural observation is that the functionals with polynomial-decay memory functions $\rho(t) = \frac{1}{t^p}, p>1, t \in [0, \infty)]$ decay slower in the asymptotic sense than the exponential memory ones. 
Hereafter, we will refer to functionals with memory functions that decay slower than the exponential decay as ``non-exponential decaying functionals''.

Notice that the Riesz representation theorem, as stated above, is specifically confined to linear functionals, which does not encompass most sequential relationships that inherently possess a nonlinear nature. 
To the best of our knowledge, a universal representation for nonlinear functionals has yet to be established.

\paragraph{Generalized memory function}

Now we introduce a memory function definition for general continuous-time sequence-to-sequence relationship $H$:
\begin{equation}\label{eq:generalized_memory_function}
    \rho(t) = \sup_{|x| \leq 1} \left | \frac{d}{dt} H_t(x \cdot \mathbf{1}_{[0, \infty)} (t) ) \right |, \qquad t \in \mathbb{R}.
\end{equation}
Notice here we only require the relationship to be smooth enough with a bounded time-derivative.
It can be verified this definition is a direct extension for the memory function $\rho(t)$ in the continuous linear functional \cref{eq:rho_in_linear_functional}. 
A memory function for discrete indices can be constructed based on the finite difference method. 
\begin{equation}\label{eq:gen_discrete_memory}
    \rho(k) = \sup_{|x| \leq 1} \left |H_{k+1}(x \cdot \mathbf{1}_{[0, \infty)} (t) )  - H_{k}(x \cdot \mathbf{1}_{[0, \infty)} (t) ) \right |, \qquad t \in \mathbb{R}.
\end{equation}
Therefore for simplicity we will use $\rho(t)$ to represent the memory function of the sequence to sequence relationship in the following sections.

\section{Main results}
\label{sec:main_results}

In this section, we first demonstrate the memory bias in mean absolute/squared error in learning linear functionals. 
Even though the linear functional represents a simplified sequence relationship, it adequately captures the memory bias phenomenon. 
Furthermore, it allows us to quantify this memory bias.

Next, the extension to temporally positive-weighted error shows the common existence of the bias. 
Only the last-term-only error is the ``unbiased'' error for linear functional. 
As reducing bias might increase the variance of the models, we propose in \cref{sec:tuning_of_tpe} to tune a specific family of temporal weight which can trade-off the bias-variance.

\subsection{Bias towards short-term memory from MAE/MSE}

Consider the sequence modelling task of approximating one-dimensional linear functional, the mean absolute/squared error are commonly used:
\begin{align}
    \textrm{Error}^{\textrm{MAE}} = \frac{1}{T} \int_0^T |y_t - \hat{y}_t| dt, \qquad \textrm{Error}^{\textrm{MSE}} = \frac{1}{T} \int_0^T (y_t - \hat{y}_t)^2 dt.
\end{align}
For simplicity, we'll drop $T$ as it's a task dependent constant. 

For linear functionals, the target output is associated to a finite time representation (see \cref{appendix:linear_functional}):
\begin{align}
    y_t = \int_0^t \rho_{t-s} x_s ds.
\end{align}
For simplicity, we first consider models without nonlinear activations such as linear RNN \citep{li2021.CurseMemoryRecurrenta}, linear temporal convolution network \citep{bai2018.EmpiricalEvaluationGeneric,jiang2023.BriefSurveyApproximation}. 
Here $\hat{\rho}$ is the model memory function defined based on \eqref{eq:generalized_memory_function}.
\begin{equation}
    \hat{y}_t = \int_0^t \hat{\rho}_{t-s} x_s ds.
\end{equation}
For linear RNN, $\hat{\rho}_s = c^\top e^{W s} U$.
For linear TCN, $\hat{\rho}_s = k_1(s) \Conv \cdots \Conv k_l(s)$. Here $k_i(s)$ is the kernel function for $i$-th convolution layer. 

Notice that the mean absolute error is minimizing the following expression
\begin{align}
    \textrm{Error}^{\textrm{MAE}}  
    & = \int_0^T \left | \int_0^t (\rho_{t-s} - \hat{\rho}_{t-s}) x_s ds \right | dt.
\end{align}
As the input sequence is uniformly distributed over a bounded set, minimizing the mean absolute error over input distribution is equivalent to minimize the following ``time-weighted memory difference'':
\begin{align}
    \mathcal{E}^{\textrm{MAE}}  & = \mathbb{E}_{\mathbf{x}} \textrm{Error}^{\textrm{MAE}}  = \mathbb{E}_{\mathbf{x}} \int_0^T \left |\int_0^t (\rho_{t-s} - \hat{\rho}_{t-s}) x_s ds \right | dt \\
    & = \int_0^T \mathbb{E}_{\mathbf{x}} \left | \int_0^t (\rho_{t-s} - \hat{\rho}_{t-s}) x_s ds \right | dt \\
    & = \int_0^T c_0 \int_0^t |\rho_{t-s} - \hat{\rho}_{t-s}| ds dt \\
    & = c_0 \int_0^T \int_0^t |\rho_{s} - \hat{\rho}_s| ds dt = c_0 \int_0^T \int_s^T |\rho_{s} - \hat{\rho}_s| dt ds \\
    & = c_0 \int_0^T (T-s) \left |\rho_{s} - \hat{\rho}_s \right | ds. \label{eq:memory_bias_of_memroy_absolute_error}
\end{align}
Here $c_0$ is some inputs sequences dependent constant.

\begin{definition}[Memory bias of mean absolute error on linear functional]
    The memory bias of mean absolute error is defined to be the weight of error on the memory function difference $|\rho_{s} - \hat{\rho}_s|$.
    According to \cref{eq:memory_bias_of_memroy_absolute_error}, the memory bias of mean absolute error on linear functional has an explicit form $b(s) = T-s$.
    This memory bias is equivalent up to a positive scaling factor $c_0$. 
    The \textit{normalized memory bias} of mean absolute error is $b(s) = \frac{2(T-s)}{T^2}$ which satisfies $\int_0^T b(s) = 1$.
\end{definition}
It can be seen the mean absolute error is biased towards learning short-term memory as it puts more weights ($b(s) = T-s$) on short-term memory error ($|\rho_{s} - \hat{\rho}_s|$). 

Based on the above derivations, we achieve the following conclusion.
\begin{theorem}
    Assume the target functional sequences $\{H_t\}_{t \in \mathbb{R}}$ are a sequence of continuous, linear, time-homogeneous, causal, regular functionals.
    The mean absolute error is biased towards short-term memory with a normalized memory bias $b(s) = \frac{2(T-s)}{T^2}$.
\end{theorem}
The definitions for functionals' continuity, linearity, time-homogeneity, causality and regularity are given in \cref{sec:definitions_for_linear_functional}.
Similar results can be derived for mean squared error by assuming the input sequences to be $L_2$-integrable.

\subsection{Temporally positive-weighted error}

Based on the above derivation, a natural idea is to adjust the temporal weight and turn the error into the following \textit{temporally positive-weighted error}. 
\begin{equation}
    \textrm{Error}^{\textrm{TPE}} = \frac{1}{T} \sum_{t=1}^T w(t) |y(t) - \hat{y}(t)|, \quad w(t) > 0.
\end{equation}
Its continuous version is $\textrm{Error}^{\textrm{TPE}} (\textrm{Continuous}) = \frac{1}{T} \int_0^T w(t) |y(t) - \hat{y}(t)| dt.$

However, the bad news is that any positive-weighted error is biased towards short-term memory for linear functionals.
\begin{theorem}\label{thm:TPE}
    Assume the target functional sequences $\{H_t\}_{t \in \mathbb{R}}$ are a sequence of continuous, linear, time-homogeneous, causal, regular functionals.
    Weight function $w: \mathbb{R}^+ \to \mathbb{R}^+$ is a positive integrable function. 
    Then the temporally positive-weighted error is biased towards short-term memory with a memory bias $b(s) = \int_{s}^T w(s) ds$.
\end{theorem}
See the proof in \cref{sec:proof_TPE}. 
Based on the memory bias' explicit form, we know to learn the linear functional without short-memory bias, the memory bias need to be a constant function. 

\begin{corollary}
    The only ``memory-unbiased'' error for continuous, linear, time-homogeneous, causal, regular functionals is the last-term-only error:
    \begin{equation}
        \textrm{Error} := |y(T) - \hat{y}(T)|.
    \end{equation}
\end{corollary}

\begin{remark}
    It is important to notice that the above unbiased result is only derived for simple linear functionals.
    For more general nonlinear functionals, it is not known whether we can have an explicit form of the memory-unbiased error. 
    Moreover, this explicit form is derived based on the assumption that there is no ``noise'' in the training data. 
    When the generalization gap is considered, the memory variance of the solution might increase as the memory bias is reduced with the last-term-only error. 
\end{remark}

\subsection{Tuning of the temporally positive-weighted errors}
\label{sec:tuning_of_tpe}

\begin{figure}[t!]
	\centering
	\includegraphics[width=0.75\textwidth]{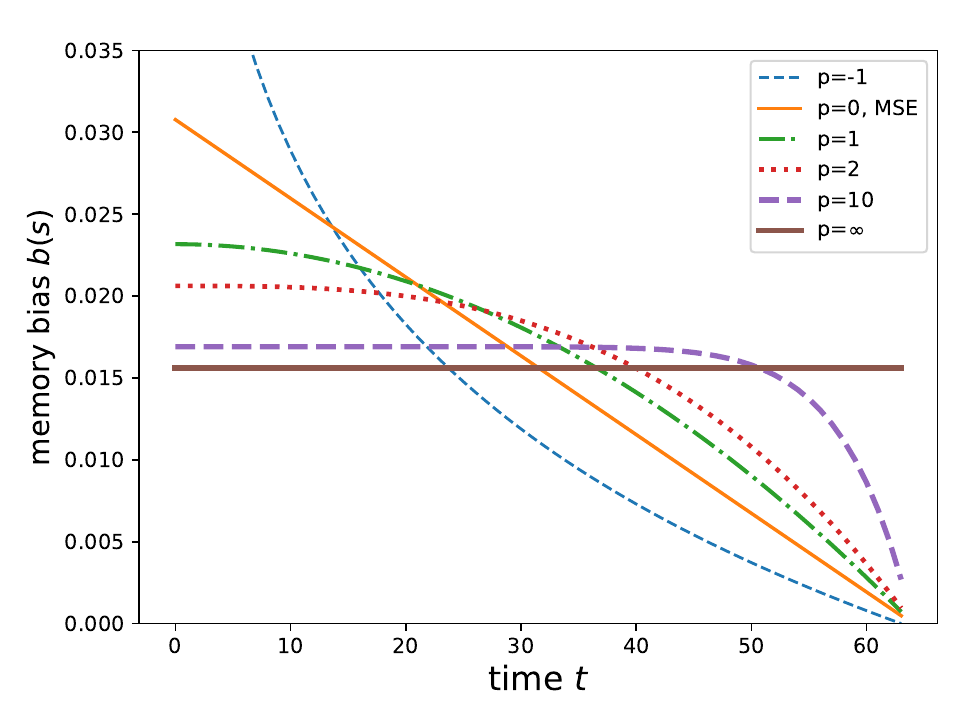}
	\caption{Normalized memory bias for different $p$}
	\label{fig:memory_bias_demo}
\end{figure}

In this subsection, we focus on a family of temporally positive-weighted error. 
The weight function is polynomial in $t$ with power $p$: $w(t) = t^p, t \geq 0$.
The corresponding memory bias for different polynomial power $p$ is evaluated and presented in \cref{fig:memory_bias_demo}.
The weight are normalized so that the total memory bias integral for each weight is 1. 
($\int_0^t b(s) ds = 1, b(s) = \int_s^T w(r) dr$.)

It can be seen that as $p$ goes to $\infty$, the memory bias flattened. 
In the limiting case, it's the unbiased last-term-only loss. 
Therefore, by properly tuning the parameter $p$ we can keep reduce the memory bias in the error. 

\begin{remark}
    Since the gradient of network is usually evaluated by $\frac{\partial Error}{\partial w} \approx \frac{\partial Error}{\partial \rho} \frac{\partial \rho}{\partial w}$. 
    It can be seen the bias in the error function will be inherited by the gradient. 
    Properly tuning the temporally weighted error can also relax the vanishing gradient issue. 
\end{remark}

\section{Numerical results}
\label{sec:numerical_results}

In this section, we use synthetic case (linear functional with polynomial decaying memory), copying problem \citep{arjovsky2016.UnitaryEvolutionRecurrenta}, and text summarization problem \citep{liu2019.TextSummarizationPretraineda} to justify the main results in the previous section. 
Notice that the numerical examples all focus on learning of long-term memory.

\subsection{Synthetic case}

We approximate the linear functional $\{ H_t(x): t \in [0, T] \}$ with linear RNN and tanh RNN. 
\begin{equation}
    H_t(x) := \int_{0}^t \rho_{t-s} x_s ds, \quad t \in [0, T].
\end{equation}
The memory function $\rho$ of the target functional is defined as $\rho(t) = \frac{1}{t^2}$, a decay function that exhibits a more gradual rate of decrease relative to the conventionally observed exponential decay in Recurrent Neural Networks.

Although the main theory is only established for linear RNN, it can be seen the result holds for general tanh RNN. 

The memory errors shown in \cref{fig:toy_monotonicity_of_p} and \cref{fig:toy_monotonicity_of_p_2} are the mean absolute error in the memory function of linear functional:
\begin{equation}
    \textrm{Memory difference} := \int_{0}^T | \rho_{s} - \hat{\rho}_{s} | dx_s.
\end{equation}
We show that as the power $p$ increased, after same number of epochs, the mean absolute memory error is smaller for larger $p$. 

\begin{table}[thb!]
    \caption{
        The memory difference is monotonic in the weight function polynomial power $p$ for different hidden dimensions (linear RNN)
        }
    \label{fig:toy_monotonicity_of_p}
    \centering
    \begin{tabular}{cccccc}
    \toprule
                        & $p=-1$ & $p=0$ & $p=1$ & $p=2$ & $p=\infty$ \\
    hidden dimension 4  & 0.167 & 0.136 & 0.120 & 0.111 & 0.085 \\
    hidden dimension 16 & 0.184 & 0.162 & 0.148 & 0.142 & 0.129 \\
    hidden dimension 64 & 0.169 & 0.103 & 0.084 & 0.074 & 0.054 \\
    \bottomrule
    \end{tabular}
\end{table}

\begin{table}[thb!]
    \caption{
        The memory difference is monotonic in the weight function polynomial power $p$ for different hidden dimensions (tanh RNN)
        }
    \label{fig:toy_monotonicity_of_p_2}
    \centering
    \begin{tabular}{cccccc}
    \toprule
                        & $p=-1$ & $p=0$ & $p=1$ & $p=2$ & $p=\infty$ \\
    hidden dimension 4  & 0.248 & 0.235 & 0.227 & 0.224 & 0.218 \\
    hidden dimension 16 & 0.322 & 0.314 & 0.307 & 0.304 & 0.299 \\
    hidden dimension 64 & 0.427 & 0.435 & 0.420 & 0.410 & 0.039 \\
    \bottomrule
    \end{tabular}
\end{table}

\subsection{Copying problem}

Apart from the synthetic linear functional task, we consider another task which is difficult for the long memory \citep{arjovsky2016.UnitaryEvolutionRecurrenta}. 
This task is trained with a different loss function cross entropy.
To show the generality of the temporally rescale error in learning long-term memory, we take temporal convolutional network \citep{bai2018.EmpiricalEvaluationGeneric} and state space model \citep{gu2021.CombiningRecurrentConvolutional, smith2023.SimplifiedStateSpace} as the models to learn the copying problem. 

\begin{figure}[tbh!]
	\centering
	\includegraphics[width=0.85\textwidth]{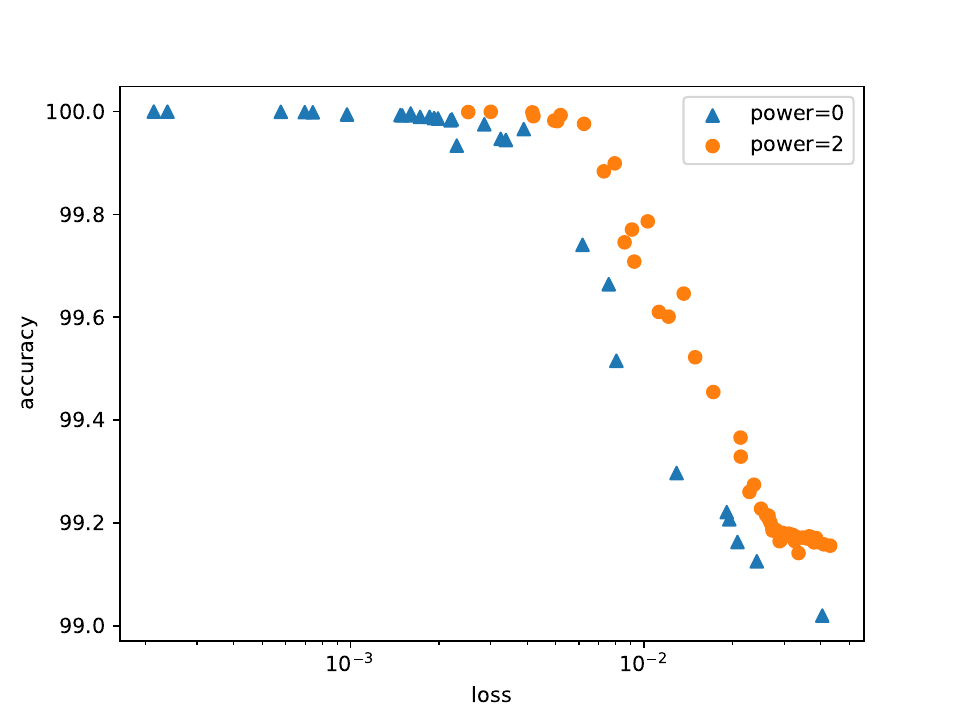}
	\caption{
        Copying Problem: Loss-Accuracy Graph for Temporal Convolutional Network. 
        The graph indicates that with the same loss value, an increase in the power $p$ leads to higher accuracy. 
        It should be highlighted that the two losses were trained with varying weights on cross-entropy, yet normalized to maintain a total weight of 1, ensuring a fair comparison.
        }
	\label{fig:copying_memory_tcn}
\end{figure}

In \cref{fig:copying_memory_tcn}, we show that larger $p$ gives a better accuracy when the loss is at the same value. 
As we rescale the error temporally but do not scale up the loss value, this is a fair comparison. 
The monotonicity in power $p$ indicates that the time-weighted loss is a more consistent error than the temporally uniform-weighted cross entropy in terms of learning long-memory task. 

\begin{figure}[htb!]
	\centering
    \includegraphics[width=0.75\textwidth]{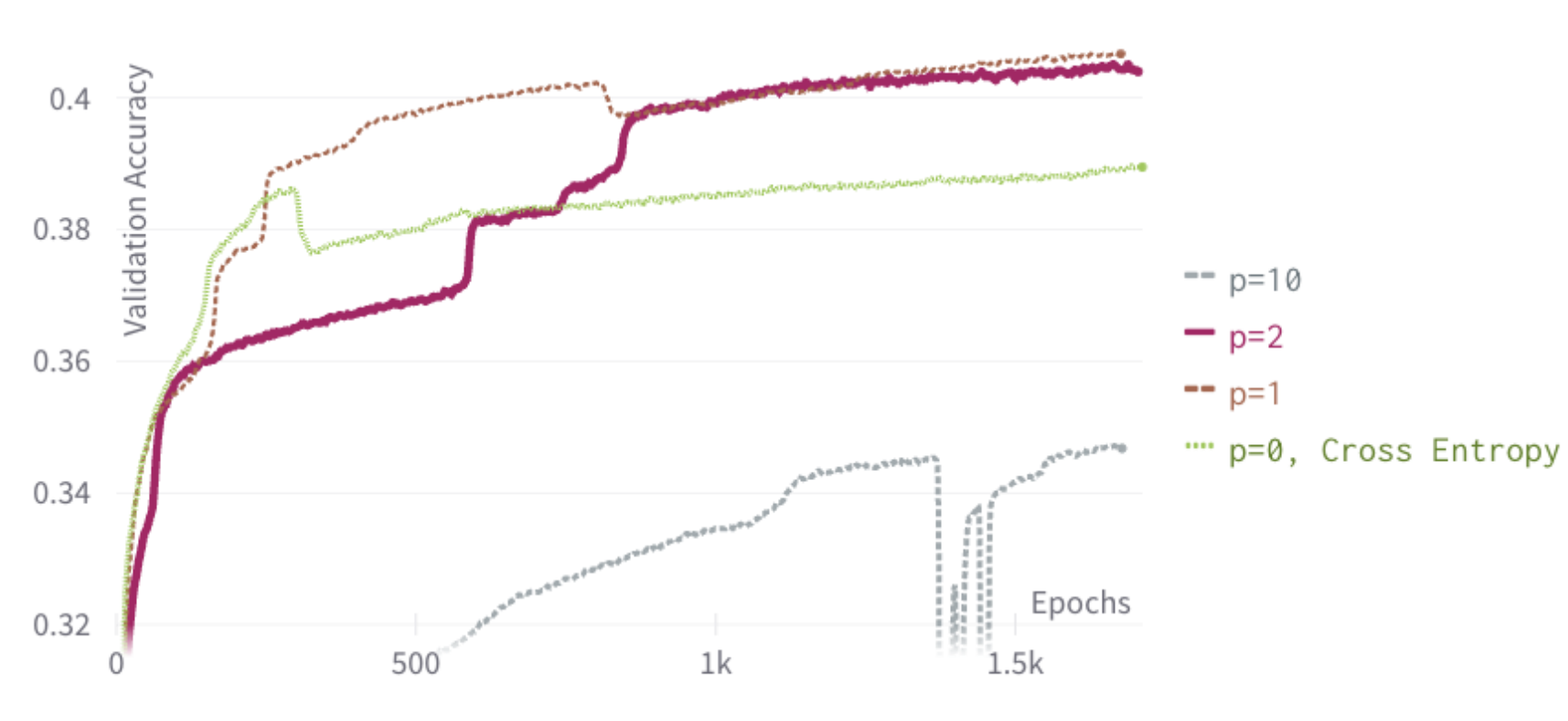}
	\caption{Copying problem, the validation accuracy for state space model (S4)}
	\label{fig:copying_memory_s4}
\end{figure}

In \cref{fig:copying_memory_s4}, we show that with sufficient training, the accuracy of larger power $p$ can further increased. 
However, as shown in the power $p=10$ case, it should be notice that learning long-term memory is in general a more difficult task, therefore learning with a larger power $p$ can be relatively slower.

\subsection{Text summarization}

In this study, we bolstered the credibility of our methods by conducting additional validation using LCSTS \citep{hu2015.LCSTSLargeScale}. 
LCSTS is an extensive compilation of Chinese text summarization datasets sourced from Sina Weibo, a prominent microblogging platform in China. 
Given the objective of generating summaries, our approach leverages long-term memory to proficiently capture and synthesize the complete content. 
For this purpose, we employed the T5-PEGASUS \citep{xue2021.MT5MassivelyMultilinguala, zhang2020.PEGASUSPretrainingExtracted} and CPT \citep{shao2022.CPTPreTrainedUnbalanced} models in our research. 
To demonstrate the effectiveness of our training, we evaluate the mean training rough-1 results after 10,000 training epochs, as shown in Table ~\ref{fig:lcsts}.

\begin{table}[thb!]
    \caption{
        The mean training rough-1 result on LCSTS after 10,000 training epochs.
        The rough-1 accuracy is monotonic in the weighted power $p$ for different models.
        }
    \label{fig:lcsts}
    \centering
    \begin{tabular}{ccccc}
    \toprule
                 & $p=0$  & $p=1$  & $p=2$  & $p=10$ \\
    T5-PEGASUS   & 0.4286 & 0.4342 & 0.4413 & 0.5212 \\
    CPT          & 0.4535 & 0.5656 & 0.5688 & 0.7354 \\
    \bottomrule
    \end{tabular}
\end{table}

\subsection{Sensitivity of \texorpdfstring{$p$}{}}

A natural question after the various numerical applications above is the sensitivity of the parameter $p$. 
Based on the main results in \cref{sec:main_results}, the best parameter $p$ for linear functional among temporally positive-weighted error is the $\infty$. 
However, as we increase $p$ in the above copying problem, it can be seen the optimization is getting increasingly difficult. 

\begin{figure}[htb!]
	\centering
	\includegraphics[width=0.75\textwidth]{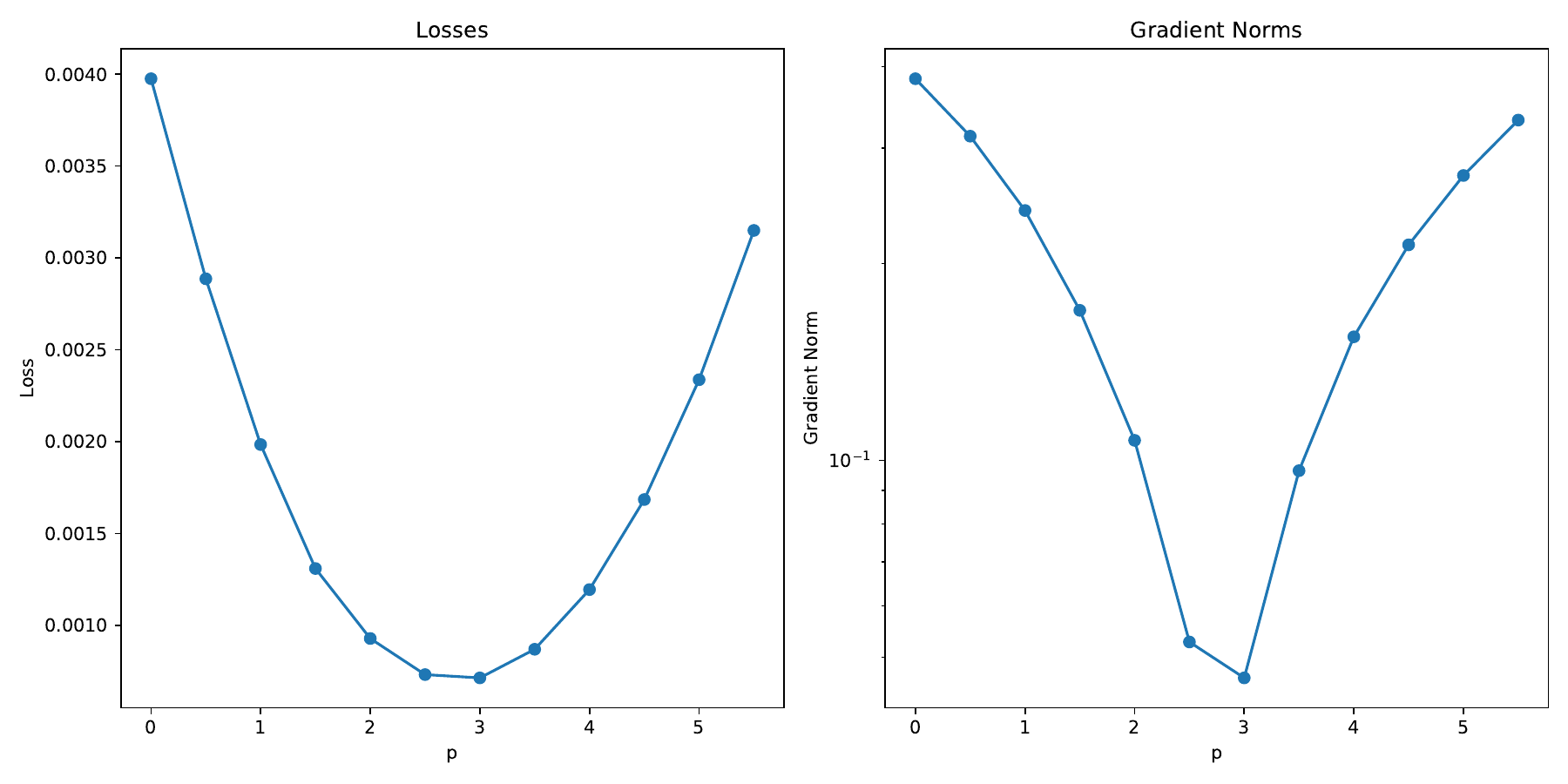}
	\caption{Sensitivity of parameter $p$ in the 1-step setting}
	\label{fig:sensitivity1}
\end{figure}

\begin{figure}[htb!]
	\centering
	\includegraphics[width=0.75\textwidth]{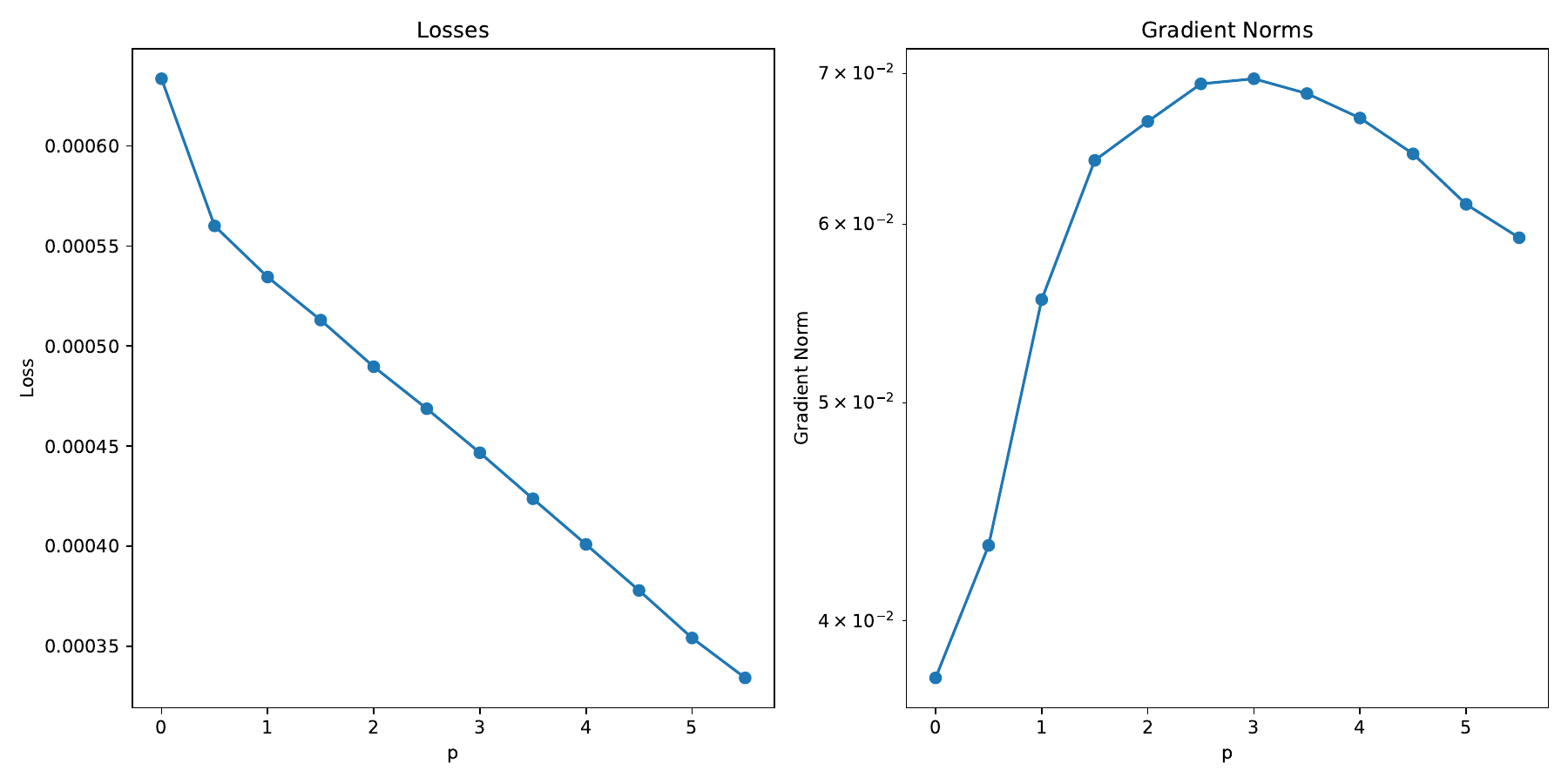}
	\caption{Sensitivity of parameter $p$ in the 16-step setting}
	\label{fig:sensitivity16}
\end{figure}

In \cref{fig:sensitivity1} and \cref{fig:sensitivity16}, we demonstrate with the linear functional task that the 1-step and 16-step results for loss decrease and gradient norm at the initialization. 
It shows that starting with a smaller $p$ usually improves the optimization. 
But to improve the learning of long-term memory, a large power $p$ is still required to reduce the memory bias.

\section{Related work}
\label{sec:related_work}

Learning long-term memory in sequence modelling is crucial, as reflected in the emergence of long short-term memory (LSTM) model \citep{hochreiter1997.LongShorttermMemory}. 
Initially proposed as a means to boost the ability of Recurrent Neural Networks (RNNs) to learn over extended periods, LSTM uses the gating mechanism to improve the long-term memory learning \citep{chen2018.DynamicalIsometryMeana}.
Similar approaches have been adopted in gated recurrent unit (GRU) \citep{cho2014.LearningPhraseRepresentations}.
The enhancement of the RNN's and LSTM's memory capabilities has been approached from a statistical perspective \citep{zhao2020.RNNLSTMHave}, leading to the proposal of setting the weight temporal decay at a polynomial rate. 
This polynomial decay corresponds to the memory function with polynomial decay in \cref{eq:generalized_memory_function}.
The LSTMp, as proposed by \citet{chien2021.SlowerBetterRevisiting}, introduces polynomial decay to the gate with the intention of improving long-term memory.

To better learn the long-term memory, it is necessary to have suitable memory metric. 
For linear functional \citep{li2022.ApproximationOptimizationTheory}, the memory function is characterized by the convolution kernel induced by the Riesz representation. 
For general nonlinear sequence to sequence relationship, there is no such representation. 
To gauge the effectiveness of differnt attempts to learn long memory, several benchmark tasks have been devised. 
A model's long-term memory ability, for instance, can be measured using a copying problem, a task specifically designed for this purpose \citep{arjovsky2016.UnitaryEvolutionRecurrenta}.
In the realm of language modelling, the LAMBADA dataset serves as a valuable tool for assessment \citep{paperno2016.LAMBADADatasetWorda}. 
A model's performance on LAMBADA often serves as an indicator of its capability to capture information from extensive and diverse contexts. 
Further, memory capacity is introduced as another crucial metric for evaluating sequence modelling \citep{gallicchio2018.ShorttermMemoryDeep}. 

From the time series perspective, statistical literature proposes the use of relative error such as mean absolute percentage error (MAPE), mean absolute scaled error (MASE) \citep{hyndman2006.AnotherLookForecast}. 
\begin{align}
    \textrm{MASE} = \frac{1}{H} \sum_{i=1}^H \frac{|y_{T+i} - \hat{y}_{T+i}|}{\frac{1}{T+H-m} \sum_{j=m+1}^{T+H} |y_j - y_{j-m}|}.
\end{align}
However, MAPE mainly takes the scaling of the sequence into consideration.
As changing the input output scale does not change the memory function scale or decay rate, the error from absolute error into relative error only removes the impact of sequence scale issue, which does not affect the definition of memory function or memory bias as they do not determine on the scale.

\section{Limitation}
\label{sec:limitation}

Although the proper temporal rescaling can improve the performance of learning long memory, this method requires the output to be sequential. 
Our method is currently limited to temporally positive-weighted error. 
The borader weight family such as nonlinear combination of relative error might further aid the learning of long-term memory.

\section{Conclusion}
\label{sec:conclusion}

In this paper, we study the bias of mean absolute/squared error towards short-term memory.
We evaluate the memory bias based on the generalized memory function, which is a natural extension of the linear functional's representation. 
Moreover, it is shown that such a memory bias exist across all temporally positive-weighted errors.
The bias problem can be relaxed by choosing a suitable weight. 
The bias-variance trade-off is discussed when the output sequence is susceptible to noise. 
Numerical experiments from synthetic dataset (linear functional) are conducted to validate the results. 
In particular, we show the discovery is not limited to synthetic case and recurrent neural networks. 
General sequence modelling such as copying problem and language modelling learned by other network structures can also benefit from the application of proper temporally positive-weighted error.

\newpage
\appendix
\onecolumn
\bibliography{main}
\newpage

\input{appendix.tex}

\end{document}

%% file: appendix.tex
\section{Linear functional and corresponding theoretical results}
\label{appendix:linear_functional}

In this section, we give the definitions of several properties for linear functional. 

\paragraph{Definitions for linear functional}
\label{sec:definitions_for_linear_functional}

\begin{enumerate}
    \item (Linearity) $H_t$ is linear if for any $c_1, c_2 \in \mathbb{R}$ and $\mathbf{x}, \mathbf{x}'\in\mathcal{X}$, $H_t(c_1 \mathbf{x} + c_2 \mathbf{x}') = c_1 H_t(\mathbf{x}) + c_2 H_t(\mathbf{x}')$.
    \item (Continuous) $H_t$ is continuous if for any $\mathbf{x},' \mathbf{x}\in \mathcal{X}$, $\lim_{{\mathbf{x}' \to \mathbf{x}}} |H_t(\mathbf{x}') - H_t(\mathbf{x})| = 0$.
    \item (Time-homogeneous) $\mathbf{H} = \{H_t: t \in \mathbb{R}\}$ is time-homogeneous if the input-output relationship interchanges with time shift: let $[S_\tau(\mathbf{x})]_t= x_{t-\tau}$ be a shift operator, then $\mathbf{H}(S_\tau \mathbf{x}) = S_\tau \mathbf{H}(\mathbf{x})$.
    \item (Causal) $H_t$ is causal if it does not depend on future values of the input. That is, if $\mathbf{x}, \mathbf{x}'$ satisfy $x_t = x_t'$ for any $t \leq T$, then $H_t(\mathbf{x}) = H_t(\mathbf{x}')$ for any $t \leq T$.
    \item (Regualrity) $H_t$ is regular if for any sequence $\{\mathbf{x}^{(k)}: k \in \mathbb{N}\}$ such that $x^{(k)}_s \to 0$ for almost every $s \in \mathbb{R}$, then $\lim_{k \to \infty} H_t(\mathbf{x}^{(k)}) = 0.$
\end{enumerate}

The linearity is a strong assumption as most of the general sequence to sequence relationship is nonlinear. 
The continuity, time-homogeneous and regularity are general properties that nice predictable sequence with temporal structures should have.

\paragraph{Riesz representation theorem}
\label{sec:Riesz_representation_theorem}

\begin{theorem}[Riesz-Markov-Kakutani representation theorem]\label{thm:riesz}
    Assume $H : \mathcal{X} \mapsto \mathbb{R}$ is a linear and continuous functional. Then there exists a unique, vector-valued, regular, countably additive signed measure $\mu$ on $\mathbb{R}$ such that
    \begin{align}
        H(\mathbf{x}) = \int_{\mathbb{R}} x_s^\top d\mu(s)
        = \sum_{i=1}^{d} \int_{\mathbb{R}} x_{s,i} d\mu_i(s).
    \end{align}
    In addition, we have $\| H \| := \sup_{\| \mathbf{x} \|_\mathcal{X} \leq 1} | H(\mathbf{x}) | = \|\mu\|_1(\mathbb{R}) := \sum_i |\mu_i|(\mathbb{R}).$
\end{theorem}

\section{Proof for \texorpdfstring{\cref{thm:TPE}}{}}
\label{sec:proof_TPE}

Consider the following weighted squared error
\begin{equation}
    \textrm{Error}^{\textrm{TPE}} = \int_0^T w(t) |y_t - \hat{y}_t| dt.
\end{equation}
It can be seen that 
\begin{align}
    \mathcal{E}^{\textrm{TPE}} & = \mathbb{E}_{\mathbf{x}} \textrm{Error}^{\textrm{TPE}} = \mathbb{E}_{\mathbf{x}}  \int_0^T w(t) \left | \int_0^t (\rho_{t-s} - \hat{\rho}_{t-s} ) x_s ds \right | dt \\
    & = c_0 \int_0^T w(t) \left (\int_0^t \left |\rho_{t-s} - \hat{\rho}_{t-s}\right | ds \right ) dt \\
    & = c_0 \int_0^T w(t) \left (\int_0^t \left |\rho_{s} - \hat{\rho}_s\right | ds \right ) dt \\
    & = c_0 \int_0^T \int_0^t w(t) \left | \rho_{s} - \hat{\rho}_s \right | ds dt \\
    & = c_0 \int_0^T \int_s^T w(t) \left | \rho_{s} - \hat{\rho}_s \right | dt ds \\
    & = c_0 \int_0^T \left ( \int_s^T w(t)dt \right )\left |\rho_{s} - \hat{\rho}_s \right | ds
\end{align}

\section{Additional numerical result}

\begin{figure}[htb!]
	\centering
    \includegraphics[width=0.75\textwidth]{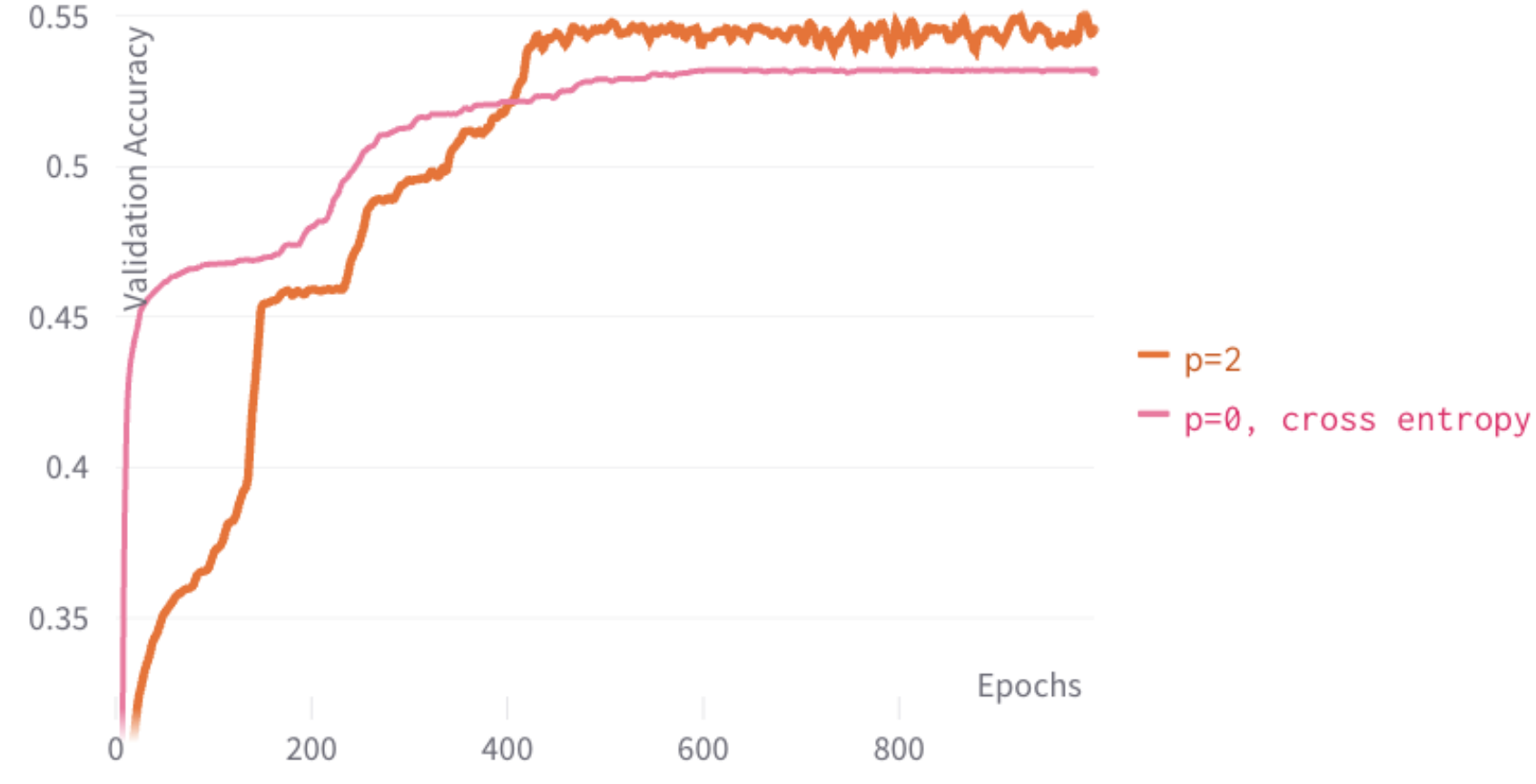}
	\caption{Copying problem, the validation accuracy for attention-based transformer}
	\label{fig:copying_memory_transformer}
\end{figure}

In addition to the temporal convolution network and state-space model (SSM), we also compare the performance of time-weighted error on attention-based transformer (see \cref{fig:copying_memory_transformer}). 
Moreover, it can be seen the gradient norm for time-weighted error with $p=2$ is larger and it relatively robust against the typical vanishing gradient issue.

\begin{figure}[htb!]
	\centering
    \includegraphics[width=0.75\textwidth]{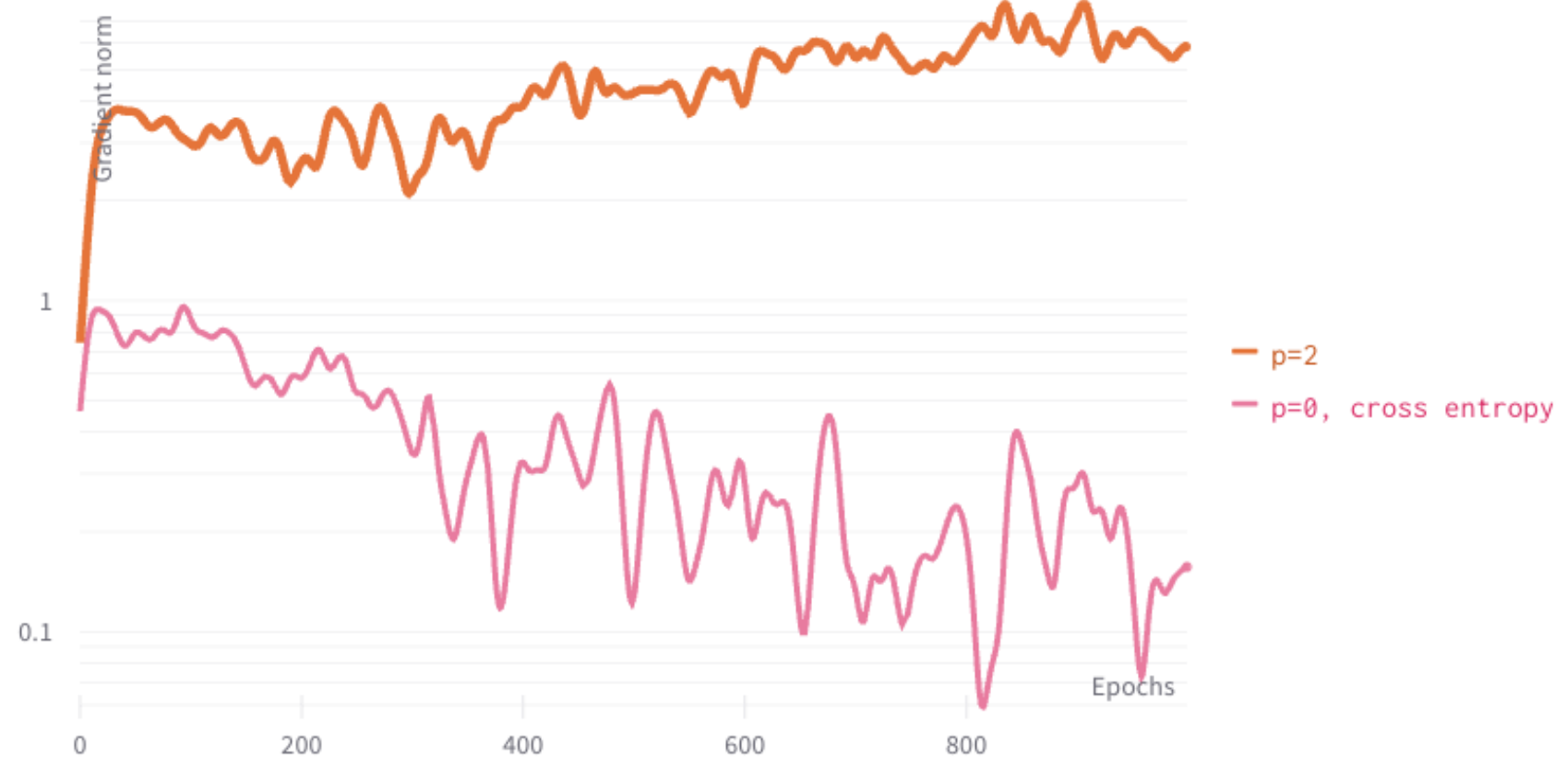}
	\caption{Copying problem, the gradient norm for attention-based transformer}
	\label{fig:copying_grad_transformer}
\end{figure}

\section{Numerical experiment details}

In this section, the setups for different numerical experiments are included. 

\subsection{Synthetic linear functional}

We use linear and tanh RNNs to learn linear functionals with polynomial decaying memory. 
The memory function of linear functional is $\displaystyle \rho(t) = \frac{1}{t^{1.1}}$.
Sequence length is 64.
Time discretization is $\Delta t = 0.1$. 
Train batch size and test batch size are 512 and 2048. 
Optimizer is Adam with learning rate 0.001. 
The training dataset is manually constructed with size 131072 while the test dataset is of size 32768. 

\subsection{Copying problem}

The copying problem is conducted based on temporal convolution network\footnote{https://github.com/locuslab/TCN} as well as state-space models\footnote{https://github.com/HazyResearch/state-spaces}.
The temporally uniform-weighted version ($w(t) \equiv 1$) is implemented based on the model provided in both repo. 
The data generation and model construction is generally the same as the example presented in the repo. 

\subsection{Text summarization}

The text summarization process relies on the utilization of an open-source repository named "t5-pegasus-pytorch"\footnote{https://github.com/renmada/t5-pegasus-pytorch}. 
We make use of the model available in this repository and implement a different optimizer, as described in Section 2. 
By leveraging the existing model and incorporating our own optimizer, we aim to enhance the performance of the text summarization process.

\subsection{Bias-variance tradeoff}

The bias-variance tradeoff is discussed based on the experiments for synthetic linear functional. 
The power $p$ is tested over $[0, \frac{1}{2}, 1, \dots, \frac{11}{2}]$.
The optimizer for 1-step and 16-step training is SGD with learning rate 0.1. 
Most of the training set is the same as synthetic linear functional.

\section{Discussion on the difficulty of training when \texorpdfstring{$p$}{} is large}

As is shown in \cref{fig:copying_memory_s4}, we observe that the accuracy of larger $p$ is increasing much slower than smaller $p$. 
We emphaisze that this result does not contradict the general rule that larger $p$ shall give less memory bias on the short term memory.
This result is verified as we prolong the training from 1500 epochs to 6000 epochs, the accuracy raises to 0.4512 which is larger than the smaller $p$. 
The higher accuracy at later stage validates the hypothesis presented in \cref{fig:sensitivity1} and \cref{fig:sensitivity16}.

\section{Computing resources}
\label{Computing-resources}

The experiments are conducted on a 20.04 Ubuntu server with 4 RTX 3090 GPUs.